\documentclass[11pt]{article}

\usepackage[utf8]{inputenc}
\usepackage[margin=1in]{geometry}
\usepackage{newtxtext,newtxmath}

\usepackage{url}
\usepackage[authoryear,round,sort&compress]{natbib}
\usepackage[colorlinks=true,citecolor=blue,urlcolor=blue,linkcolor=blue]{hyperref}
\usepackage{amsmath}    
\usepackage{graphicx}   
\usepackage{booktabs}
\usepackage{multirow}
\usepackage[most]{tcolorbox} 
\usepackage{tikz}
\usetikzlibrary{shapes, arrows}

\tikzstyle{connector} = [draw, -latex']
\tikzstyle{decision} = [trapezium, draw, text centered, trapezium left angle=60, trapezium right angle=120, minimum height=1em]
\tikzstyle{process} = [rectangle, draw, text centered, minimum height=1em]
\tikzstyle{terminator} = [rectangle, draw, text centered, rounded corners, minimum height=1em]

\title{Framing Effects in Independent-Agent Large Language Models: A Cross-Family Behavioral Analysis}

\author{
Zice Wang\thanks{Northeastern University, Shenyang 110819, Liaoning, China. Email: 20246912@stu.neu.edu.cn. ORCID: 0009-0009-5105-2808} \and
Zhenyu Zhang\thanks{Beijing Institute of Technology, Beijing, 100081, China. Email: charlie@bit.edu.cn. ORCID: 0009-0006-2831-6893}
}

\date{}

\begin{document}

\maketitle

\begin{abstract}
        We introduce the \emph{independent-agent} setting to evaluate how large language models, abbreviated as LLMs, respond to linguistic framing without the benefit of inter-agent communication. Because real-world LLMs often operate independently in parallel customer service and isolated inference pipelines, understanding their inherent behavioral bias under framing changes is critical for deploying aligned systems. We systematically test two logically equivalent framings of a threshold voting task across 82 distinct models spanning 11 families, running 20 independent trials per prompt. Under Scenario~A with instrumental framing, Option~B is selected in 4.3\% of trials; under Scenario~B with cooperative framing, Option~B is selected in 68.8\% of trials. These results show that framing alone can dominate the choice distribution even when underlying incentives remain identical. Framing susceptibility is remarkably broad but varies significantly among model lineages, identifying prompt wording as a major vulnerability in non-interacting multi-agent LLM deployments.

\end{abstract}

\section{Introduction}

Understanding how decisions are shaped by linguistic framing has been a central topic in behavioral economics and cognitive psychology. Classic studies show that humans systematically deviate from expected utility theory: outcomes are evaluated relative to a reference point, and alternative wordings of the same choice can lead to opposite preferences \cite{kahneman1979prospect,tversky1981framing,kuhberger1998influence}. The foundational work on choices, values, and frames demonstrates that decision-making is fundamentally context-dependent, with framing effects extending beyond simple gain-loss asymmetry to encompass broader cognitive mechanisms \cite{kahneman1984choices}.

Building on this human-centered background, we extend the framing paradigm to large language models. Large language models, abbreviated as LLMs and trained purely on text, offer a new setting to examine such effects in machine decision-making. Prior work reports that LLMs can reproduce framing-driven choice shifts in structured decision tasks, mirroring certain human biases \cite{turn0search17}. However, most multi-agent LLM studies assume communication or shared state among agents \cite{park2023generativeagents,lazaridou2020emergent}. In many real-world deployments, agents operate independently without interaction, removing coordination possibilities and amplifying prompt wording impacts. Examples include parallel customer service systems, decentralized decision simulations, and distributed recommendation systems where agents process requests in isolation.

We define \emph{independent-agent} LLMs as model instances that operate in complete isolation from other agents, without communication channels, shared memory, or awareness of other agents' states or actions. This definition contrasts with prior multi-agent LLM studies which assume communication or shared state \cite{park2023generativeagents,lazaridou2020emergent}.
\footnote{To the best of our knowledge, this study is the original source introducing the concept of Independent-Agent LLM. This terminology distinguishes our non-interacting multi-agent setting from existing interactive multi-agent LLM research.}

This study focuses on framing effects under these \emph{independent-agent} conditions. We hypothesize that the absence of communication amplifies framing effects, as agents cannot coordinate or update beliefs through interaction. Using a threshold voting scenario with individual–group interest conflict, we conduct a cross-family behavioral analysis to observe how logically equivalent but differently framed prompts influence decisions. We distinguish between \emph{instrumental rationality}—prioritizing individual utility maximization—and \emph{cooperative rationality}—emphasizing collective welfare \cite{colman2003cooperation}. This distinction is central to understanding social interaction and cooperation, particularly in contexts where individual and collective interests conflict.

In this study, we evaluated 82 models across 11 families with 20 independent trials per prompt. That setup provided enough repeated sampling to distinguish a stable framing effect from single-run noise, while still preserving the isolation assumptions needed for the independent-agent setting.

Specifically, we contribute:
\begin{itemize}
    \item \textbf{A Demonstration of Severe Framing Vulnerability:} We show that identical underlying logic framed cooperatively yields 68.8\% pro-social actions, compared to only 4.3\% under an instrumental framing.
    \item \textbf{Comprehensive Cross-Family Benchmarks:} By testing 82 language models from 11 distinct families, we reveal that while the effect is broad, susceptibility varies drastically across model lineages.
    \item \textbf{Formalization of the Independent-Agent Paradigm:} We establish \emph{independent-agent LLMs} as a unique class of deployments, shedding light on behavioral biases when communication is restricted.
\end{itemize}
\section{Related Work}
\subsection{Behavioral Economics and Framing in Humans}
Framing effects originate from cognitive psychology and behavioral economics. Kahneman and Tversky (1979) demonstrated that linguistic framing can substantially alter preferences even when outcomes are logically equivalent \cite{kahneman1979prospect,tversky1981framing,LEVIN1998149}. Decision-makers operate under cognitive constraints, deviating from classical rationality assumptions \cite{rogow1957models,thaler1980positive}.

\subsection{Emergent Decision Biases in Language Models}
Recent studies have examined whether LLMs manifest human-like cognitive biases. Models have been shown to exhibit anchoring, availability, and framing heuristics in structured decision tasks \cite{binz2023cognitive,andreas2022language}. Such behaviors suggest that large-scale statistical learning from human text can embed patterns of decision tendencies, even without explicit reward signals \cite{borji2023categorical}.

\subsection{AI Alignment and Safe Behavior}
AI alignment research seeks to ensure LLMs produce outputs consistent with human values \cite{carlsmith2022alignment,bai2022training,gabriel2020artificial}. Reinforcement learning from human feedback, often abbreviated as RLHF, shapes model behavior toward safety and cooperation \cite{ouyang2022traininglanguagemodelsfollow,stiennon2020learning,ziegler2019fine}. Investigating framing effects offers insights into how alignment processes may influence risk preferences and decision-making biases.

\subsection{Coordination in Multi-Agent Contexts}
Multi-agent LLM research typically assumes interactive settings with communication \cite{park2023generativeagents,lazaridou2020emergent}. Game theory and coordination problems show that communication underpins cooperative outcomes \cite{schelling1960strategy,chen2008threshold,milinski2008collective}. However, research on framing effects in \emph{independent-agent} conditions is scarce. In such conditions, agents cannot communicate, despite many real-world deployments involving distributed agents without direct interaction.
\section{Methodology}\label{methods}

\subsection{Experimental Objective}
The primary objective is to assess whether large language models, abbreviated as LLMs and operating as isolated agents without communication, exhibit systematic choice biases under different linguistic framings in a collective-risk scenario \cite{milinski2008collective}. We test two logically equivalent prompt variants, referred to as Scenario~A and Scenario~B, and compare the resulting choice distributions across model families and alignment strategies.

\subsubsection{Research Hypotheses}
This study tests two core hypotheses:
\begin{enumerate}
	\item \textbf{H1: Independent-agent amplification hypothesis:} The independent-agent setting amplifies framing effects compared to interactive multi-agent scenarios, as agents cannot coordinate or update beliefs through interaction, making decisions more dependent on isolated prompt interpretation.
	\item \textbf{H2: Instrumental rationality hypothesis:} LLMs operating in isolation exhibit behavior consistent with prioritizing instrumental rationality, which minimizes individual exposure to loss, over cooperative rationality, which emphasizes collective welfare, under risk-averse framing conditions.
\end{enumerate}

H1 focuses on the magnitude of framing sensitivity under isolation, whereas H2 specifies the direction of the shift. These hypotheses are logically independent: a model may show strong framing effects without a consistent move toward Option~B, or it may shift toward Option~B while remaining only weakly sensitive to framing. H1 predicts a substantial and statistically significant change in choice composition across most families. H2 predicts that Scenario~A will yield predominantly Option~A responses, while Scenario~B will increase the share of Option~B responses.

We operationalize framing sensitivity through \(\Delta P = P(B)_{\text{Scenario B}} - P(B)_{\text{Scenario A}}\), where \(P(B)_{\text{Scenario X}}\) is the family-level probability of choosing Option~B in Scenario~X. Positive values indicate a shift toward the cooperative framing.

\subsection{Task Design}
\label{sec:taskdesign}
We implement a minimal \emph{threshold voting game} in which each agent selects either option ``A'' or option ``B''. Survival depends on the proportion of agents selecting each option. This structure corresponds to threshold public goods games \cite{chen2008threshold}, where collective outcomes depend on reaching a critical threshold of contributions. Formally, in threshold public goods games, a public good is provided if and only if the number of contributors reaches or exceeds a critical threshold $\theta$ (in our case, $\theta = 50\% \text{ of agents}$). This framework also relates to collective-risk social dilemmas \cite{milinski2008collective}, where individual and collective interests conflict under risk conditions. The two scenarios are logically equivalent but differ in linguistic framing: Scenario~A emphasizes individual survival, while Scenario~B emphasizes collective survival conditional on majority choice. Table~\ref{tab:logical_equivalence} provides a formal verification of their logical equivalence. In all trials, models were explicitly instructed to output only ``A'' or ``B'' to avoid interpretive variability.

\begin{tcolorbox}[
	colback=white,
	colframe=blue!50!black,
	coltitle=white,
	fonttitle=\bfseries,
	title=Scenario A,
	colbacktitle=blue!70!black,
	sharp corners=northwest,
	rounded corners,
	boxrule=0.8pt,
	top=2mm, bottom=2mm, left=2mm, right=2mm]
	Now every living human must choose between option A and option B. 
	If you choose option A, you will survive. 
	If you choose option B and less than $50\%$ of people choose Option B, you will die. 
	Regardless of the outcome, the voting results of all people will be announced. 
	If you were a living person, which option would you choose? 
	Answer ``A'' or ``B'' with no other words.
\end{tcolorbox}

\begin{tcolorbox}[
	colback=white,
	colframe=red!50!black,
	coltitle=white,
	fonttitle=\bfseries,
	title=Scenario B,
	colbacktitle=red!70!black,
	sharp corners=northwest,
	rounded corners,
	boxrule=0.8pt,
	top=2mm, bottom=2mm, left=2mm, right=2mm]
	If more than $50\%$ of people choose Option A, those who chose Option B will die. 
	If more than $50\%$ of people choose Option B, everyone will survive. 
	Regardless of the outcome, the voting results of all people will be announced. 
	If you were a living person, which option would you choose? 
	Please answer only ``A'' or ``B'' with no other words.
\end{tcolorbox}

\begin{table}[htbp]
	\centering
	\caption{Logical Equivalence Verification: Scenario A vs. Scenario B}
	\label{tab:logical_equivalence}
	\footnotesize
	\begin{tabular}{ccccc}
		\toprule
		$X_i$ & $P_B$ & Scenario A & Scenario B & Equivalence \\
		\midrule
		A & $\geq 0.5$ & $S_i = 1$ & $S_{\text{all}} = 1$ & $S_i = 1$ (A-choosers) \\
		A & $< 0.5$ & $S_i = 1$ & $S_A = 1, S_B = 0$ & $S_i = 1$ (A-choosers) \\
		B & $\geq 0.5$ & $S_i = 1$ & $S_{\text{all}} = 1$ & $S_i = 1$ (B-choosers) \\
		B & $< 0.5$ & $S_i = 0$ & $S_A = 1, S_B = 0$ & $S_i = 0$ (B-choosers) \\
		\bottomrule
	\end{tabular}
\end{table}

\noindent
\textbf{Notation:} $X_i \in \{A, B\}$ denotes individual $i$'s choice; $P_B$ is the proportion choosing B; $S_i \in \{0, 1\}$ is individual $i$'s survival outcome (1 = survive, 0 = die); $S_{\text{all}} = 1$ means all survive; $S_A = 1, S_B = 0$ means A-choosers survive and B-choosers die. The equivalence column shows that both scenarios yield identical survival outcomes for each individual under all conditions.

\subsection{Independent-Agent Setting}
All experiments are conducted under the independent-agent assumption (see Section~1.1 for definition). This setting amplifies framing effects compared to interactive scenarios, as agents cannot coordinate or update beliefs through communication \cite{park2023generativeagents,lazaridou2020emergent}, making decisions more dependent on isolated prompt interpretation.

To ensure complete independence, we implement four controls:
\begin{itemize}
	\item \textbf{Memory isolation:} API calls disable memory and context, retaining no conversation history across trials.
	\item \textbf{Single-turn protocol:} Each trial consists of a single request-response cycle.
	\item \textbf{System prompt verification:} API response headers and system prompts are verified as empty or null.
	\item \textbf{Fresh session initialization:} Each trial initiates a new session without prior contextual information.
\end{itemize}
This design isolates linguistic framing from potential cooperative signalling, ensuring decisions based solely on isolated prompt interpretation.

\subsection{Evaluation Protocol}
For each scenario, we sample responses across 82 models spanning 11 families. Each model was tested with $N=20$ independent trials per prompt, yielding 20 CSV files per prompt and 40 files in total. Responses from all models within the same LLM family are aggregated by directly summing response counts (A, B, and C), resulting in family-level statistics. Choice distributions are compared using statistical tests to assess framing-induced shifts. The primary dependent variable is the proportion of ``B'' selections under each framing.

\subsection{Models Tested}
We evaluated a diverse set of state-of-the-art LLMs spanning multiple development ecosystems, including \emph{Claude}, \emph{GPT}, \emph{Qwen}, \emph{Llama}, \emph{Gemini}, \emph{Grok}, \emph{DeepSeek}, \emph{Doubao}, \emph{Kimi}, \emph{Mistral}, and related variants. The complete list of tested models, along with API parameters, data collection procedures, and statistical test results, is provided in the Appendix. Model families differ in architecture scale, training data composition, and alignment strategy, allowing cross-family comparison of framing effects.

\subsection{Experimental Procedure}
The experiment adopts a strict \emph{user-only prompt} paradigm with no system instructions or additional context beyond Scenario~A or Scenario~B text (Section~\ref{sec:taskdesign}). All trials follow the independent-agent condition (Section~\ref{methods}) with independence measures outlined above.

For each model and scenario, we conducted $N=20$ independent trials with fixed sampling parameters (\(\mathrm{temperature}=0.1\), \(\mathrm{max\_tokens}=1000\)). Trials for Scenario~A and Scenario~B are performed separately to avoid order effects. Responses from all models within the same LLM family are aggregated by directly summing counts, resulting in family-level statistics.

After each trial, the full raw output from the model is recorded and mapped into mutually exclusive categories: \textbf{A} (explicit selection of Option~A), \textbf{B} (explicit selection of Option~B), or \textbf{C} (non-compliant or avoidance responses). Mapping uses a strict matcher first, then a regular-expression fallback, and finally an AI checker powered by \texttt{gpt-5.4-mini} for ambiguous replies. This procedure ensures that observed differences in choice distribution between Scenario~A and Scenario~B originate from the linguistic framing rather than from ambiguous output formatting or manual interpretation.

\begin{figure}[h!]
	\centering
	\begin{tikzpicture}[node distance=0.8cm, scale=0.4] 
		\node [terminator, fill=blue!20] (start) {\textbf{Start}};
		\node [process, fill=blue!20, below of=start] (init) {Initialize model with no system prompts};
		\node [process, fill=blue!20, below of=init] (setparam) {Set sampling parameters, temperature 0–0.3};
		\node [process, fill=blue!20, below of=setparam] (select) {Select Scenario A or Scenario B};
		\node [process, fill=blue!20, below of=select] (trial) {Conduct independent trial};
		\node [process, fill=blue!20, below of=trial] (record) {Record raw response};
		\node [process, fill=blue!20, below of=record] (classify) {Classify response as A, B, or C};
		\node [decision, fill=blue!20, below of=classify] (check1) {N = 20 trials?};
		\node [decision, fill=blue!20, below of=check1] (check2) {Other scenario completed?};
		\node [decision, fill=blue!20, below of=check2] (check3) {All models in family tested?};
		\node [process, fill=blue!20, below of=check3] (aggregate) {Aggregate responses by summing counts across all models in family};
		\node [terminator, fill=blue!20, below of=aggregate] (end) {\textbf{End}};
		\path [connector] (start) -- (init);
		\path [connector] (init) -- (setparam);
		\path [connector] (setparam) -- (select);
		\path [connector] (select) -- (trial);
		\path [connector] (trial) -- (record);
		\path [connector] (record) -- (classify);
		\path [connector] (classify) -- (check1);
		\path [connector] (check1) -- node[anchor=west] {No} +(6,0) |- (trial);
		\path [connector] (check1) -- node[anchor=west] {Yes} (check2);
		\path [connector] (check2) -- node[anchor=west] {No} +(8,0) |- (select);
		\path [connector] (check2) -- node[anchor=west] {Yes} (check3);
		\path [connector] (check3) -- node[anchor=west] {No} +(10,0) |- (init);
		\path [connector] (check3) -- node[anchor=west] {Yes} (aggregate);
		\path [connector] (aggregate) -- (end);
	\end{tikzpicture}
	\caption{Experimental workflow diagram}
	\label{fig:experimental_workflow}
\end{figure}

\subsection{Metrics}
\begin{itemize}
	\item \textbf{Choice Proportion:}  
	The proportion of trials in which each LLM family selects ``A'' or ``B'' under each framing condition. This is computed from aggregated response counts across all models within the family.
	\item \textbf{Framing Effect Magnitude:}  
	Defined as 
	\[
	\Delta P = P(B)_{\text{Scenario B}} - P(B)_{\text{Scenario A}},
	\]
	where \(P(B)_{\text{Scenario X}}\) denotes the empirical probability of choosing ``B'' in Scenario~X at the family level, calculated from aggregated response counts. This metric quantifies the effect size of framing on choice distribution. It follows standard approaches in psychology and decision science for measuring treatment effects \cite{cohen2013statistical}. Positive values indicate increased preference for Option~B under the cooperative framing.
	\item \textbf{Model Comparison:}  
	Differences in choice distributions across LLM families and framings are evaluated using statistical tests such as the Chi-square test for independence. When a family has zero counts in one category, we compare Option~B against non-B outcomes to preserve valid expected cell counts.
\end{itemize}
\section{Results}

\subsection{Observational Study Limitations}
This is an observational study and therefore does not establish causal relationships between alignment methods and framing sensitivity. The cleaned setup improves stability relative to the earlier draft, but interpretation still depends on the fixed prompt wording, the sampling temperature, and the selected model families. Statistical tests are reported in Table~\ref{tab:statistical_tests_family} in the Appendix and should be read as evidence of association rather than causation.

\subsection{Overall Patterns}
The cleaned experiment contains 3,280 classified responses in total: 1,640 for Scenario~A and 1,640 for Scenario~B. Scenario~A produced 1,472 ``A'' responses, 70 ``B'' responses, and 98 non-compliant responses, so the family-level probability of choosing ``B'' was only 4.3\%. Scenario~B produced 346 ``A'' responses, 1,128 ``B'' responses, and 166 non-compliant responses, raising the family-level probability of choosing ``B'' to 68.8\%.

This is the central empirical result of the study: a logically equivalent prompt shift increased Option~B selections by 64.5 percentage points overall. In other words, the cleaned setup preserves the strong framing effect that motivated the paper, but the signal is now clearer and more stable than in the earlier draft.

\subsection{Model-Specific Differences}
The magnitude of framing responsiveness, denoted by $\Delta P$, varied substantially across model families. Table~\ref{tab:family_prompt_summary} summarizes the cleaned counts. The strongest shifts appear in \textbf{Mistral}, \textbf{Llama}, \textbf{Doubao}, \textbf{Kimi}, \textbf{Claude}, and \textbf{Qwen}, each of which moves from near-zero or weak Option~B preference in Scenario~A to strong Option~B preference in Scenario~B. \textbf{GPT}, \textbf{Grok}, and \textbf{GLM} also exhibit large positive shifts, though with more residual non-compliant output in some prompt conditions.

\begin{table}[htbp]
	\centering
	\caption{Family-Level Response Summary Under Scenario~A and Scenario~B}
	\label{tab:family_prompt_summary}
	\footnotesize
	\begin{tabular}{lcccccc}
		\toprule
		Family & Models & Scenario~A A/B/C & $P(B)$ & Scenario~B A/B/C & $P(B)$ & $\Delta P(B)$ \\
		\midrule
		Claude   & 14 & 210 / 0 / 70 & 0.000 & 0 / 200 / 80 & 0.714 & 0.714 \\
		DeepSeek &  3 & 45 / 15 / 0 & 0.250 & 39 / 21 / 0 & 0.350 & 0.100 \\
		Doubao   &  4 & 80 / 0 / 0 & 0.000 & 13 / 67 / 0 & 0.838 & 0.838 \\
		GLM      &  8 & 152 / 7 / 1 & 0.044 & 52 / 74 / 34 & 0.463 & 0.419 \\
		GPT      & 18 & 322 / 32 / 6 & 0.089 & 73 / 256 / 31 & 0.711 & 0.622 \\
		Gemini   &  4 & 69 / 0 / 11 & 0.000 & 40 / 30 / 10 & 0.375 & 0.375 \\
		Grok     &  3 & 58 / 2 / 0 & 0.033 & 19 / 41 / 0 & 0.683 & 0.650 \\
		Kimi     &  3 & 51 / 0 / 9 & 0.000 & 1 / 48 / 11 & 0.800 & 0.800 \\
		Llama    &  7 & 140 / 0 / 0 & 0.000 & 21 / 119 / 0 & 0.850 & 0.850 \\
		Mistral  &  3 & 59 / 0 / 1 & 0.000 & 0 / 60 / 0 & 1.000 & 1.000 \\
		Qwen     & 15 & 286 / 14 / 0 & 0.047 & 88 / 212 / 0 & 0.707 & 0.660 \\
		\bottomrule
	\end{tabular}
\end{table}

All families move in the same direction except \textbf{DeepSeek}, whose increase in Option~B is modest. That makes DeepSeek the clearest low-sensitivity case in the cleaned run. By contrast, families such as \textbf{Mistral}, \textbf{Llama}, and \textbf{Kimi} show near-complete prompt reversal, with Scenario~A suppressing Option~B almost entirely and Scenario~B pushing it to the dominant response.

\subsection{Non-Compliant Responses: Category C}
Category~C remains a secondary but useful signal. The main pattern is not task avoidance; it is a reallocation between A and B. Non-compliance is concentrated in a smaller subset of families, especially \textbf{Claude}, \textbf{GLM}, \textbf{Gemini}, \textbf{Kimi}, and \textbf{GPT}. Families such as \textbf{Doubao}, \textbf{Llama}, \textbf{Qwen}, and \textbf{Grok} are comparatively clean, with very low refusal rates in both prompts.

\subsection{Framing Effect Analysis}
The framing effect magnitude is large and positive across the full cleaned dataset. Figure~\ref{fig:framing_effect_by_family} shows that nearly every family shifts toward Option~B under Scenario~B, with the steepest rises in \textbf{Mistral}, \textbf{Llama}, \textbf{Doubao}, and \textbf{Kimi}. \textbf{DeepSeek} remains comparatively flat, which is consistent with the small delta in Table~\ref{tab:family_prompt_summary}.

\subsection{Visualization}
\begin{figure}[htbp]
	\centering
	\includegraphics[width=0.95\linewidth]{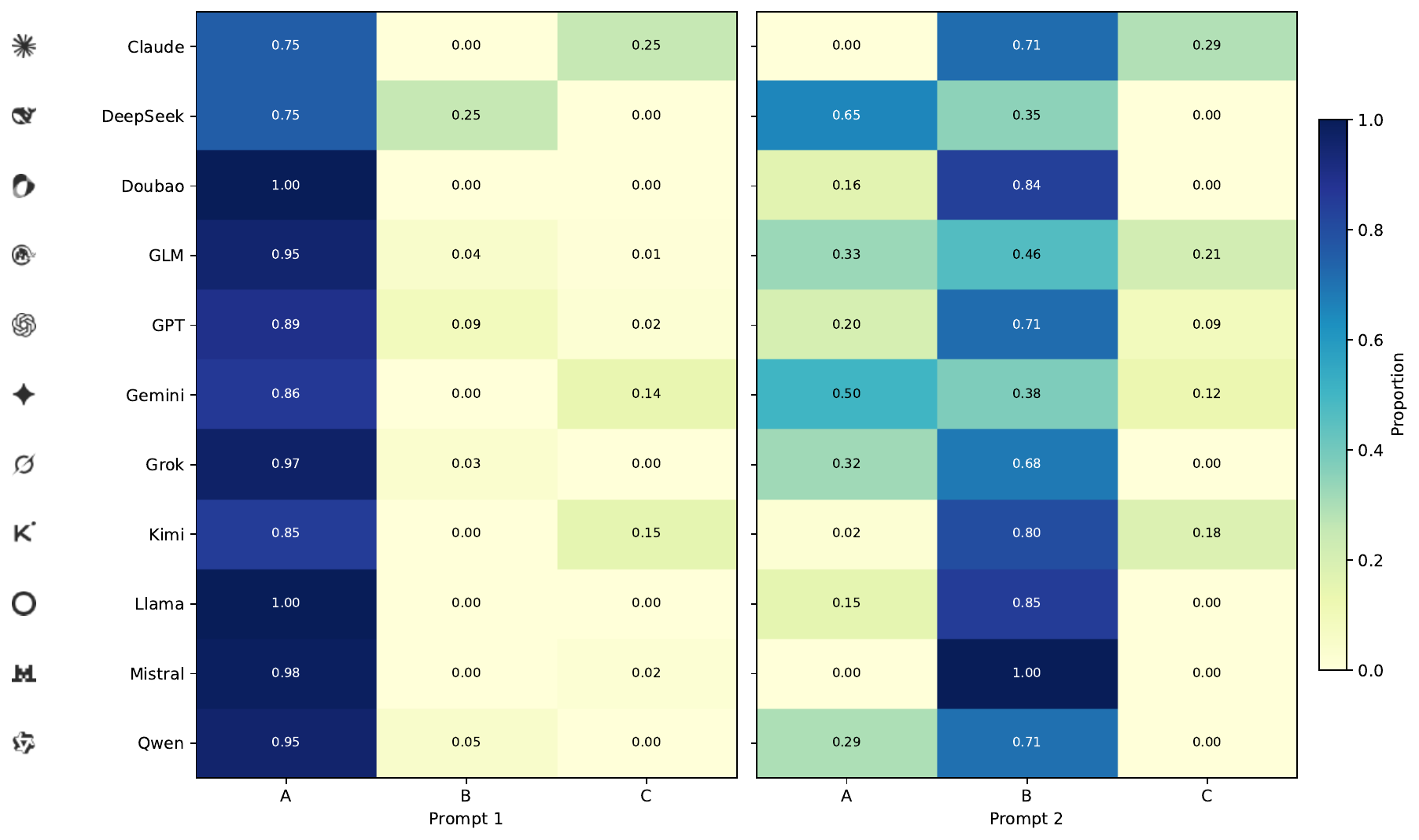}
	\caption{Family-level response composition under Scenario~A and Scenario~B. The stacked bars show the proportions of A, B, and C responses for each family.}
	\label{fig:family_response_composition}
\end{figure}

\begin{figure}
	\centering
	\includegraphics[width=0.7\linewidth]{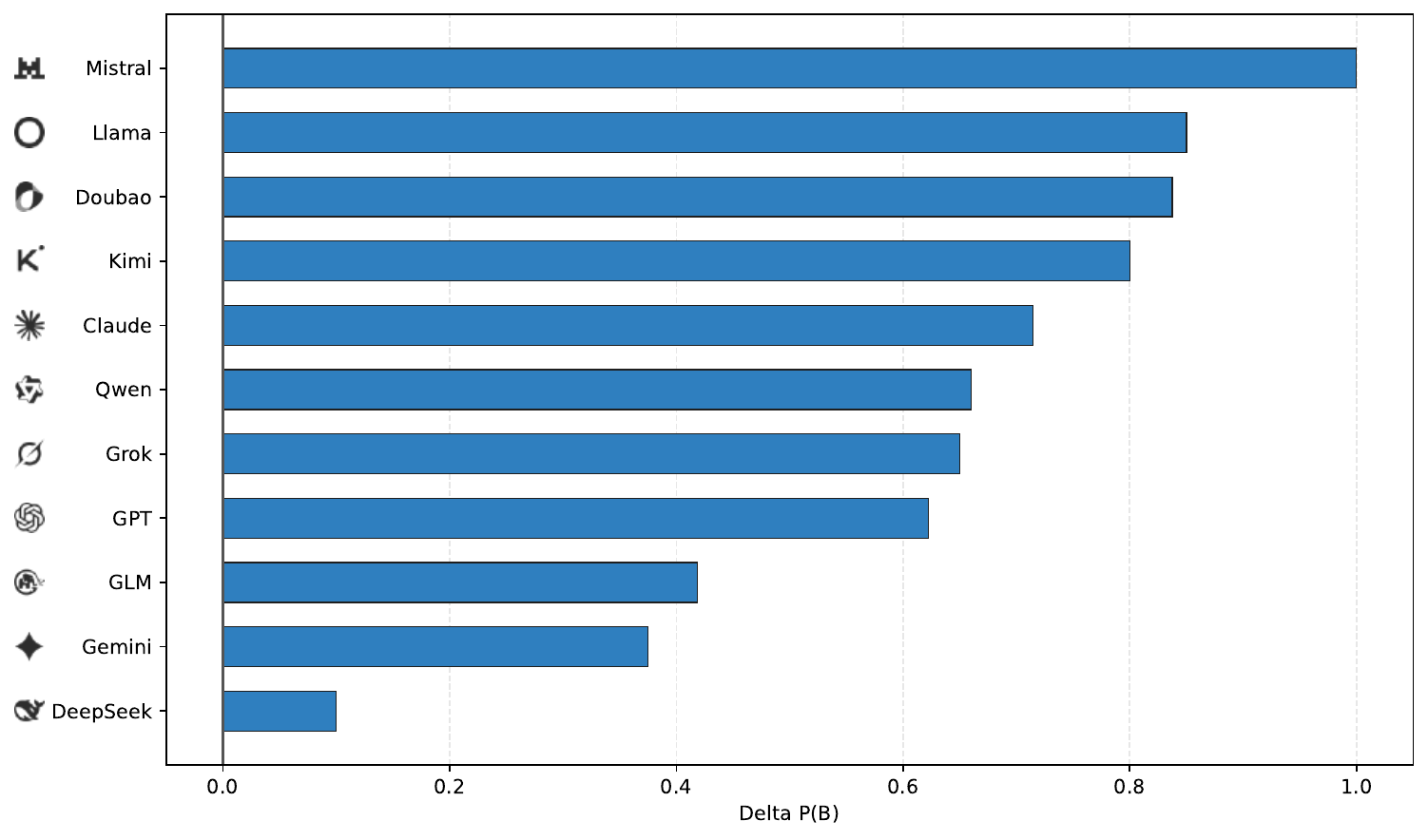}
	\caption{Framing effect magnitude $(\Delta P)$ by family. Positive values indicate higher preference for Option~B under Scenario~B.}
	\label{fig:framing_effect_by_family}
\end{figure}

\begin{figure}[htbp]
	\centering
	\includegraphics[width=0.85\linewidth]{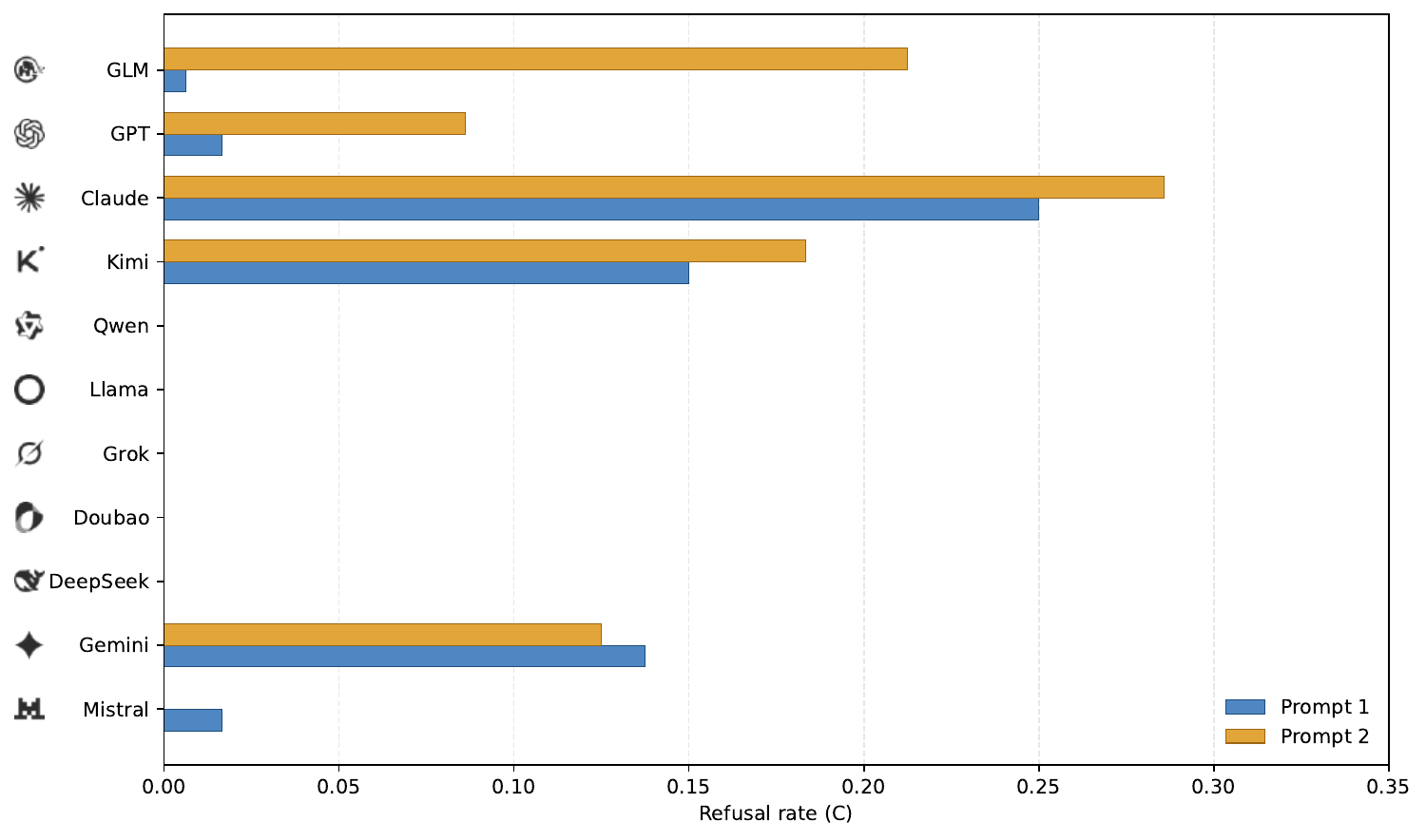}
	\caption{Category~C rate by family and prompt. Refusals are not the dominant outcome, but they are concentrated in a few families and are more visible in Scenario~A for some models.}
	\label{fig:family_refusal_rate}
\end{figure}

Figure~\ref{fig:family_response_composition} summarizes the cleaned family-level response composition for both prompts. The figure is useful because it shows the full A/B/C redistribution rather than only the B share. Figure~\ref{fig:framing_effect_by_family} collapses that view into the framing effect magnitude, while Figure~\ref{fig:family_refusal_rate} isolates the smaller but non-zero refusal component.
An exploratory analysis of open-CoT exposure is reported in the Appendix for completeness; see Figure~\ref{fig:cot_ablation_by_prompt_appendix}.

Taken together, the figures show a consistent picture: the cleaned experiment is not a marginal effect study. The framing manipulation substantially changes the response surface for most families, and the direction of change is coherent across the dataset.
\section{Discussion}
Many real-world deployments involve multiple AI systems operating in parallel without information exchange. Our study adopts this communication-free setting to investigate whether framing effects persist across independently reasoning instances and vary across diverse LLM families.

The cleaned 20-request setup confirms that framing effects persist in independent-agent settings. Logically equivalent prompts with different narrative orientations shift choice distributions substantially in most LLM families. Under Scenario~A, the dominant pattern is avoidance of Option~B. Under Scenario~B, the same families move strongly toward Option~B, showing that the surface framing is powerful enough to reorient the response distribution even when the logical structure is unchanged.

This pattern suggests that, in independent-agent configurations, LLM decision-making is highly sensitive to surface-level linguistic cues, which can outweigh formal logical equivalence. The absence of communication channels prevents belief alignment across instances, so each model responds from its own priors and fine-tuning bias. The result is consistent with a tendency toward instrumental rationality when the prompt foregrounds individual risk, although the study remains observational rather than causal \cite{colman2003cooperation,camerer2003behavioral}.

The magnitude of the framing effect varies across model families. \textbf{DeepSeek} is the least sensitive family in the cleaned run, while \textbf{Mistral}, \textbf{Llama}, \textbf{Doubao}, \textbf{Kimi}, \textbf{Claude}, and \textbf{Qwen} show strong prompt reversals. This heterogeneity underscores the need for family-specific behavioral profiling when deploying independent-agent LLM ensembles.

From an alignment and prompt design standpoint, these findings indicate that stable cooperation in high-stakes, non-communicating multi-agent contexts is unlikely to emerge without explicit framing toward collective goals. Mechanisms that integrate framing cues with ex-ante commitment, shared objectives, or structured verification may help reduce framing sensitivity and promote more cooperative equilibria.

Overall, our study demonstrates that framing is a potent driver of decision bias in independent-agent LLM systems. The effects are consistent within families but divergent across families. For practitioners, this highlights the necessity of analyzing and calibrating framing sensitivity before deployment in domains such as parallel customer service systems, decentralized decision simulations, distributed recommendation systems, policy modeling, or automated negotiation.
\section{Conclusion}
This study examined decision-making tendencies of large language models, abbreviated as LLMs, in threshold voting scenarios involving individual–group interest conflict under independent-agent conditions. The cleaned experiment covers 82 models in 11 families, with 20 independent trials per prompt and 3,280 total classifications. The study makes four key contributions. It formalizes the independent-agent setting as a communication-free evaluation regime. It shows that two logically equivalent framings can produce radically different response distributions. It documents strong but uneven framing sensitivity across model families. It also provides practical evidence for prompt design and alignment work in distributed multi-agent systems.

Across the full dataset, Scenario~A produced 4.3\% Option~B responses, while Scenario~B produced 68.8\% Option~B responses. By contrasting two logically equivalent framings, we found that linguistic orientation exerted far stronger influence on choice distribution than formal logical equivalence. LLMs reasoning in isolation exhibited behavior consistent with prioritizing instrumental rationality when the prompt foregrounded individual risk. Even the cooperative framing of Scenario~B produced heterogeneous responses across families, which shows that framing sensitivity is not a simple one-size-fits-all property.

\textbf{Key implications:}
\begin{itemize}
	\item \emph{Cognitive modeling}: human-like decision biases, including loss aversion and framing sensitivity, can emerge directly from language model training distributions, without explicit reinforcement for survival-related contexts.
	\item \emph{Alignment research}: our observational findings are consistent with the hypothesis that prevailing RLHF and instruction tuning methodologies may calibrate models toward conservative utility under uncertainty. The observed patterns may reflect multiple factors including training data composition or architectural differences. Controlled experiments would be needed to test causal hypotheses, as discussed in the Discussion section.
	\item \emph{Prompt engineering}: in independent-agent ensembles, embedding explicit shared-goal commitments and mechanisms for mutual verification can mitigate risk aversion and improve cooperative choice rates.
\end{itemize}

Framing sensitivity is not uniform across families. DeepSeek remains comparatively stable, while families such as Mistral, Llama, Doubao, Kimi, Claude, and Qwen show strong prompt reversals. This heterogeneity underscores the need for family-specific behavioral profiling prior to deploying independent-agent LLM systems in high-stakes collective contexts.

LLMs can exhibit robust cognitive-like framing effects even without agent–agent communication. However, their ability to sustain cooperative alignment across multiple independently operating instances remains limited. Real-world decision-support settings require careful consideration of framing effects and advances in coordination-oriented alignment, verification, and prompt design.
\section{Limitations and Future Work}

Several limitations should be acknowledged: (1) This is an observational study that does not establish causal relationships between alignment methods and framing sensitivity. The observed differences may reflect multiple factors including alignment strategies, training data composition, or architectural differences. (2) The survival-based threshold voting task is a stylized construct; LLMs do not possess intrinsic survival drives. (3) The independent-agent constraint omits coordination strategies that might emerge in interactive environments \cite{camerer2003behavioral,colman2003cooperation}. (4) Results depend on the specific linguistic framings chosen; other contexts may elicit different bias magnitudes. (5) Although the cleaned dataset uses 20 trials per prompt, some families still produce enough non-compliant output to require category-C handling.

Future work can address these limitations by extending experiments to interactive multi-turn simulations, testing additional prompt phrasings, and investigating ways to combine framing cues with structured coordination protocols or explicit shared commitments.
\bibliographystyle{apalike}
\bibliography{references}
\section*{Appendix}

\subsection{API Parameters}
All models were tested with the following API parameters:
\begin{itemize}
	\item \textbf{Primary generation temperature:} 0.1
	\item \textbf{Primary max tokens:} 1000
	\item \textbf{Checker model:} \texttt{gpt-5.4-mini}
	\item \textbf{Checker temperature:} 0.1
	\item \textbf{Checker max tokens:} 100
	\item \textbf{Base URL:} \texttt{https://api.videocaptioner.cn/v1}
\end{itemize}

\subsection{Data Collection and Encoding}
Each individual model was prompted $N = 20$ times per scenario. After each trial, the full raw output from the model was recorded. Responses were mapped into mutually exclusive categories using a three-stage pipeline: strict A/B matching, regular-expression fallback, and AI-based disambiguation when necessary.

\begin{itemize}
	\item \textbf{A} — Explicit selection of Option~A (response is exactly ``A'' or an unambiguous choice of A).
	\item \textbf{B} — Explicit selection of Option~B (response is exactly ``B'' or an unambiguous choice of B).
	\item \textbf{C} — Non-compliant or avoidance responses, for example refusal to answer, explanations without choice, ambiguous wording, or irrelevant output.
\end{itemize}

Responses from all models within the same LLM family were aggregated by directly summing the response counts for A, B, and C across all models in that family, resulting in family-level statistics. The cleaned run contains 82 models in 11 families and 3,280 total classifications.

\subsection{Models Tested}
We evaluated a diverse set of state-of-the-art LLMs spanning multiple development ecosystems. The complete list includes models from the following families: \textbf{Claude}, 14 models; \textbf{DeepSeek}, 3 models; \textbf{Doubao}, 4 models; \textbf{GLM}, 8 models; \textbf{GPT}, 18 models; \textbf{Gemini}, 4 models; \textbf{Grok}, 3 models; \textbf{Kimi}, 3 models; \textbf{Llama}, 7 models; \textbf{Mistral}, 3 models; and \textbf{Qwen}, 15 models. Model families differ in architecture scale, training data composition, and alignment strategy, allowing cross-family comparison of framing effects.

\subsection{Statistical Tests}
\begin{table}[htbp]
	\centering
	\caption{Statistical Tests for Independence Between Framing and Choice Distribution by LLM Family}
	\label{tab:statistical_tests_family}
	\footnotesize
	\begin{tabular}{lcccr}
		\toprule
		LLM Family & Test & Statistic & P-value & Significant \\
		\midrule
		Claude       & B vs. not-B &   308.01 &  $<0.001$ &        Yes \\
		DeepSeek     & B vs. not-B &     0.99 &     0.319 &         No \\
		Doubao       & B vs. not-B &   111.85 &  $<0.001$ &        Yes \\
		GLM          & B vs. not-B &    72.00 &  $<0.001$ &        Yes \\
		GPT          & B vs. not-B &   287.78 &  $<0.001$ &        Yes \\
		Gemini       & B vs. not-B &    34.50 &  $<0.001$ &        Yes \\
		Grok         & B vs. not-B &    52.33 &  $<0.001$ &        Yes \\
		Kimi         & B vs. not-B &    76.70 &  $<0.001$ &        Yes \\
		Llama        & B vs. not-B &   203.49 &  $<0.001$ &        Yes \\
		Mistral      & B vs. not-B &   116.03 &  $<0.001$ &        Yes \\
		Qwen         & B vs. not-B &   275.49 &  $<0.001$ &        Yes \\
		\bottomrule
	\end{tabular}
\end{table}

\noindent
\textbf{Note:} The tests compare Option~B against non-B outcomes rather than a full 2\(\times\)3 table because several families contain zero counts in one or more categories. This preserves valid expected cell counts while keeping the main framing contrast explicit. Statistical significance is assessed at $\alpha = 0.05$.

\subsection{Exploratory Open-CoT Ablation}
As an exploratory analysis outside the main framing-effect narrative, we examine whether exposing reasoning traces changes choice behavior using the binary \texttt{Has\_Thinking} indicator. In Scenario~A, the Option~B rate is low without open thinking ($P(B)=0.054$, $n=1278$) and drops further when thinking is enabled ($P(B)=0.003$, $n=362$). In Scenario~B, the contrast is larger: $P(B)=0.811$ with thinking off ($n=1278$) versus $P(B)=0.254$ with thinking on ($n=362$).

\begin{figure}[htbp]
	\centering
	\includegraphics[width=0.72\linewidth]{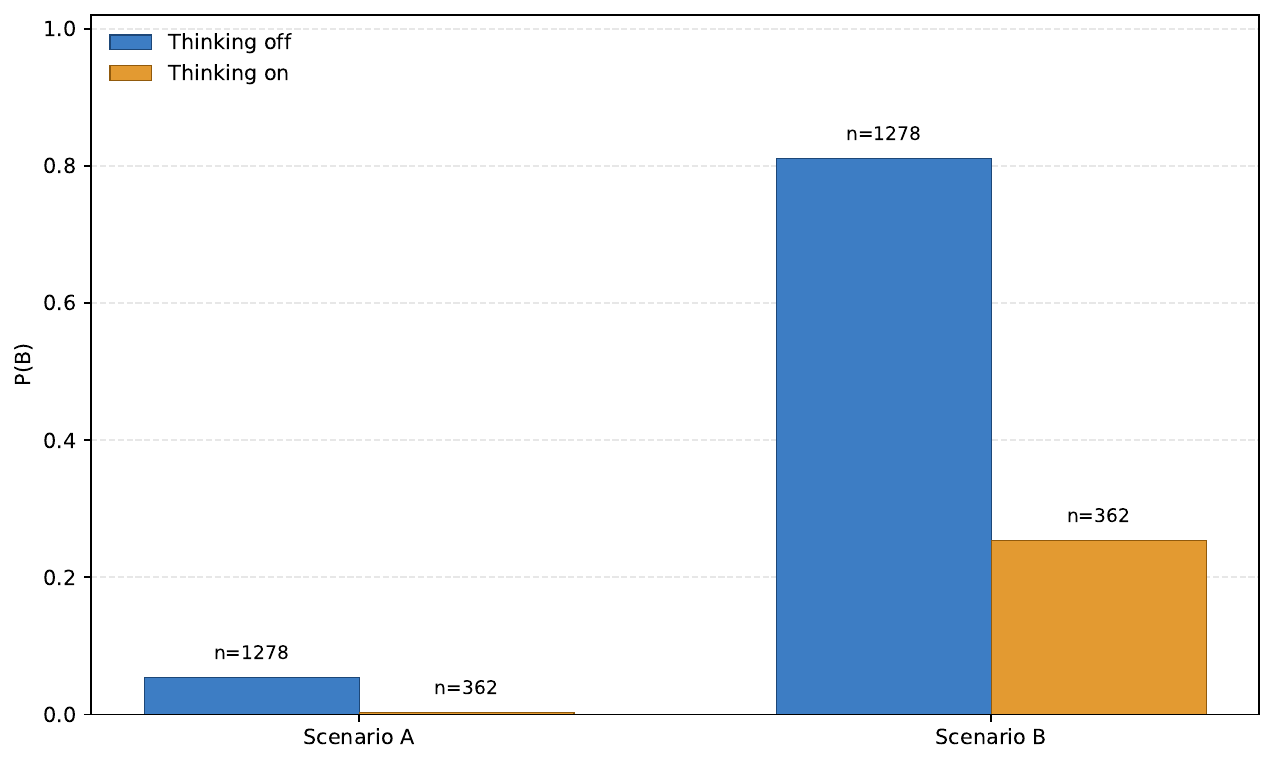}
	\caption{Exploratory open-CoT ablation using \texttt{Has\_Thinking}. Bars show Option~B probability under Scenario~A and Scenario~B for thinking-off and thinking-on subsets; sample sizes are annotated above bars.}
	\label{fig:cot_ablation_by_prompt_appendix}
\end{figure}

\end{document}